\documentclass{article}

     \usepackage[nonatbib]{neurips_2019}
\usepackage{graphicx}
\usepackage{amsfonts}
\usepackage[utf8]{inputenc} 
\usepackage[T1]{fontenc}    
\usepackage{hyperref}       
\usepackage{url}            
\usepackage{booktabs}       
\usepackage{amsfonts}       
\usepackage{nicefrac}       
\usepackage{microtype}      
\title{Neural Architecture Search for Joint Optimization of Predictive Power and Biological Knowledge}

\author{%
	Zijun Zhang \\
	Department of Statistics, UCLA \\
	\texttt{zj.z@ucla.edu}
	\And
	Linqi Zhou \\
	Department of Statistics, UCLA \\
	\texttt{alexzhou907@gmail.com }
	\AND
	Liangke Gou \\
	Department of Statistics, UCLA \\
	\texttt{lgou@ucla.edu}
	\And
	Ying Nian Wu \\
	Department of Statistics, UCLA \\
	\texttt{ywu@stat.ucla.edu}
}

\begin{document}

\maketitle

\begin{abstract}
We report a neural architecture search framework, BioNAS, that is tailored for biomedical researchers to easily build, evaluate, and uncover novel knowledge from interpretable deep learning models. The introduction of knowledge dissimilarity functions in BioNAS enables the joint optimization of predictive power and biological knowledge through searching architectures in a model space. By optimizing the consistency with existing knowledge, we demonstrate that BioNAS optimal models reveal novel knowledge in both simulated data and in real data of functional genomics. BioNAS provides a useful tool for domain experts to inject their prior belief into automated machine learning and therefore making deep learning easily accessible to practitioners. BioNAS is available at \url{https://github.com/zj-zhang/BioNAS-pub}.

\end{abstract}


\section{Introduction}
Deep learning has been successfully applied to many genomics and biomedicine problems \cite{Eraslan2019}. The modern genomic studies employ deep learning to build predictive models based on large scale data, such as high-throughput sequencing of DNA \cite{Zhou2015,Kelley2016} or RNA \cite{Xiong2015,Zhang2019}. Deep learning has also been adopted to biomedical image analysis for diagnostics \cite{Esteva2017}.

Despite many successful implementations, an Achilles heel for deep learning in bioinformatics applications has been its black-box nature \cite{Ching2018}. When and how the practitioners could trust (and not trust) a model's prediction is vital, as the prediction tasks are directly or indirectly linked to medicine or patient treatments. Hence, building interpretable deep learning models and understanding the decision logics in highly predictive models is an urgent call in biomedicine \cite{Yu2018}.

With the continued cost reduction in high-throughput sequencing, the genomics field is arguably one of the largest contributors for today's big data era \cite{Eraslan2019}. Predictive deep learning models based on genomic DNA and RNA sequences have shed new lights on discovering the molecular regulatory patterns as well as the therapeutics potentials in various human diseases \cite{Zhou2015,Xiong2015,Zhou2019}. However, harnessing deep learning requires non-trivial amount of parameter tuning and code development, hence it imposes a computational barrier for life science domain experts. Building an automated machine learning framework that meets the specific needs of domain researchers can democratize big data science, and potentially speed up the feedback loop of data generation.

\begin{figure}[t]
	\begin{center}
		\includegraphics[scale=0.3]{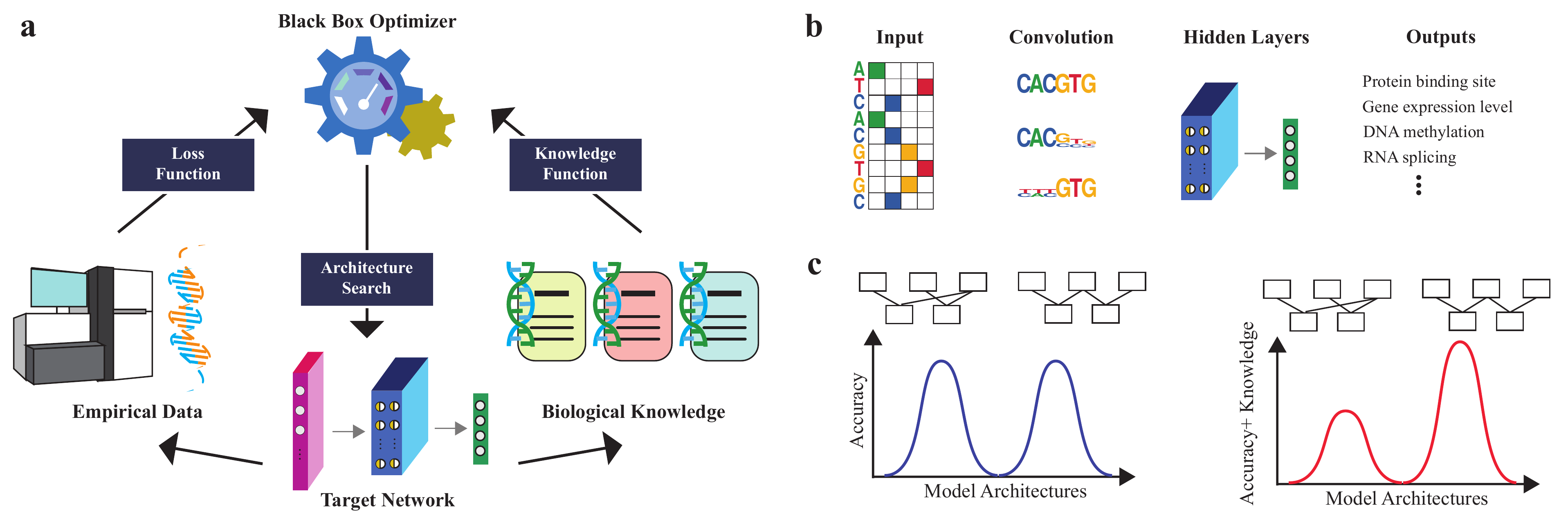}
		\caption{An overview of the BioNAS framework. \textbf{(a)} The generic neural architecture framework for joint optimization of predictive power on empirical data and knowledge consistency with existing biological knowledge. \textbf{(b)} General workflow for applying convolutional neural networks to genomic sequence tasks. Sequences are one-hot encoded, then fed into convolution, pooling and fully-connected layers, to perform a variety of predictive tasks. First-layer convolutional feature maps are often enriched in biological motifs. \textbf{(c)} A hypothetical example of two architectures with similar predictive power. After consideration of knowledge consistency, one architecture compares favorably to the other.}
		\label{fig1}
	\end{center}
\end{figure}

In this work, we propose a neural architecture search framework, BioNAS, that is tailored for life science researchers in genomic sequence analysis using deep learning (Figure \ref{fig1}a). BioNAS incorporates biological knowledge through a generic family of knowledge dissimilarity functions, to balance the interpretability and predictive power in architecture searching. 

We used BioNAS to search convolutional neural network architectures with genomic sequences as inputs. The generic tasks using genomic sequences as inputs for convolutional neural networks are illustrated in Figure \ref{fig1}b. With the fast development of high-throughput sequencing machines, genomic sequences have become an important source of information for elucidating the regulatory patterns in computational biology and functional genomics. The genomic sequences are first encoded as one-hot matrices in general. Then the convolutional neural networks take the one-hot encoded genomic sequences as inputs and provides an end-to-end computational framework for a wide variety of predictive tasks, including DNA/RNA sequence properties such as protein-DNA/RNA interaction sites and methylation levels, and gene activities such as expression and splicing levels. In particular, the first layer of convolutional filters are feature extractors and are often enriched in biological motifs \cite{Zhou2015,Kelley2016}. Matching motifs to the convolutional filters improved the interpretability of the CNNs and advanced the understanding of regulatory patterns in many biological sequence applications \cite{Zhou2015,Kelley2016}, while also provided useful approaches for transfer learning to related tasks \cite{Zhou2019}. However, the influence of different neural network architectures on the interpretability of the first-layer convolutional filters remains elusive. Methods for robustly improving the model interpretability while maintaining prediction accuracy are absent.

We demonstrate the necessity of considering knowledge dissimilarity for searching interpretable models, and the efficiency of finding both predictive and interpretable models by BioNAS. Finally, novel knowledge discovery from trained deep learning models is facilitated by optimizing uncovered prior knowledge, as we show in both simulated data and in real data of functional genomics.

\section{Related Work}

\textbf{Deep learning using genomic sequences}: The application of deep learning to solve biological questions was pioneered by three methods: DeepBind \cite{Alipanahi2015}, DeepSEA \cite{Zhou2015} and Basset \cite{Kelley2016}. DeepBind trained multiple single-task models, while DeepSEA and Basset employed a multitask learning framework. All three methods used raw genomic DNA sequence data as inputs to predict specific DNA properties. With the rapid accumulation of sequencing data, there have been numerous efforts utilizing convolutional neural networks to predict various DNA, RNA and gene-level molecular phenotypes \cite{Eraslan2019}.

\textbf{Decoding deep learning for biomedicine}: Recent efforts of opening the black-box of deep learning in biomedicine can be cataloged in two types. First, methods are developed to understand the decision rules after models are trained. This includes the back propagation-based methods \cite{Shrikumar2017}, and permutation-based methods \cite{Zhou2015}. More broadly, ideas and implementations for interpreting generic deep learning models, e.g. approximating the non-linear decision boundaries locally with linear classifiers \cite{Ribeiro2016}, are in principle applicable to genomics models as well. In essence, these methods are "model decoders" in the BioNAS framework that decode trained model to a human-understandable space (Figure \ref{fig1}; also see Methods). These methods can be readily supplemented to the naive model decoder discussed in this paper, depending on particular applications. 

A second class of methods aim to obtain interpretable models through customized tensor operations during training. This includes separable fully-connected layer (SFC) \cite{Alexandari2017} and de novo layer \cite{Ploenzke2018}. While outperforming built-in layers in the reported tasks \cite{Ploenzke2018,Alexandari2017}, these layers might not be suitable for all data types and tasks, hence posing a challenge for domain experts to fully utilize their advantages. Depending on different data and contexts, these customized layers can be incorporated in the target model space and subsequently evaluated by BioNAS in a data-driven manner.

\textbf{Hyperparameter optimization}: Various methods have been developed for hyperparameter optimization. A popular class of methods are based on Bayesian Optimization, which has been shown to outperform random search or grid search through the consideration of the past history of target function evaluations \cite{Kandasamy2019, smac-2017}. A sub-domain for hyperparameter optimization that is more specialized for deep learning is called Neural Architecture Search (NAS). NAS was first proposed and implemented through reinforcement learning \cite{Zoph2016}. While reinforcement learning and Bayesian optimization aim to optimize an unknown black-box function, another method that directly parameterized the architecture and used gradients to search layer connectivity was recently proposed and discussed \cite{Liu2018}. BioNAS can employ any optimizer for searching the neural architecture so long as the optimizer works with the loss function and the knowledge dissimilarity function.

\section{Methods}

\subsection{Knowledge dissimilarity function design}

We propose a family of generic functions, i.e. knowledge dissimilarity function $K(\cdot, \cdot)$, to measure the distances between a set trained model parameters denoted as $W$, and some existing knowledge $A$. The knowledge $A$ can be information from orthogonal experiments conducted on the same experimental materials, similar experiments conducted in another relevant biological system, or through textbooks and previous research articles. Because of the diverse forms of $A$ and the its inherent incompleteness and uncertainties, we introduce another encoder function $h(A)$ to transform abstract knowledge into a probabilistic representation. On the other hand, extracting relationships from trained model weights is non-trivial, a model decoder function $f(W)$ is designated to distill the model decision rules. Collectively, the knowledge dissimilarity function is dissected into two components, which transform abstract knowledge and model weights into the same space to be compared.

In this work we aim to design knowledge dissimilarity functions for Convolutional Neural Network (CNN) models with genomic sequences as inputs. The input genomic sequences of $N$ nucleotides are one-hot encoded into a $N\times4$ matrix (Figure \ref{fig1}b). Genomic sequences have short and conserved patterns, namely motifs \cite{Dhaeseleer2006}. Motifs scatter sparsely and are presumed with biological functions in the genome. Previous experimental and statistical approaches have established the probabilistic representations of these motifs, called positional weight matrices \cite{Dhaeseleer2006}.

Specifically, the knowledge $A$ is a set of foreground DNA sequences generated by biological experiments that are presumably enriched in functional motifs. By contrasting the foreground sequences to a set of random background DNA sequences, statistical methods based on mixture modeling \cite{Bailey1994,Zhou2004} can effectively detect the motif positional weight matrices. These motif-learning methods are the knowledge encoder function $h(A)$ in this application. The output of $h(A)$ is a $n\times4$ matrix, with the motif length being $n$, and each row following a multinomial distribution with respect to four DNA letters. We start with the simplest case where a single motif is encoded as knowledge. Let $\tilde{W} \in \mathbb{R}^{n\times4}$ denote the $h(A)$ encoded knowledge for a given motif.

To understand the learned patterns in the CNN weights, we look at the first convolutional layer with $J$ filters, with the $j$-th filter denoted as $W^j, \forall j \in \{1,2,..,J\}$. Each filter weight $W^j$ is a $m\times4$ matrix, i.e. $W^j \in \mathbb{R}^{m\times4}$, where $m$ denotes the filter width, and 4 columns correspond to four DNA letters. Let $w^j_{ik}, \forall i\in\{1,2,..,m\}, \forall k\in\{1,2,3,4\}$ be the element of $i$-th row and $k$-th column in the $j$-th convolutional filter. 

In order to compare a convolutional filter $W^j$ to the encoded knowledge $\tilde{W}$, we apply a softmax function $f(W^j)$ on each row $W^j_{i}=[w^j_{i1}, w^j_{i2}, w^j_{i3}, w^j_{i4}]$ by 

\begin{equation}
\hat{W^j_i}=f(W^j_i)=\frac{\exp(\beta w^j_{ik})}{\sum_{k} \exp(\beta w^j_{ik}) }
\end{equation}
where $\beta$ is a pre-defined positive value representing temperature. In this paper we set $\beta=0.1$.

The output of $f(W^j)$ is still a $m\times4$ matrix with each row following a multinomial distribution; we let $\hat{W}:=f(W)=\{f(W^j)\in \mathbb{R}^{m\times4}, \forall j\}$ be the decoded model weights.

Now with the encoded knowledge $\tilde{W}:=h(A)$ and the decoded model weights $\hat{W}:=f(W)$, it's straight-forward to measure the dissimilarity function $K(\cdot, \cdot)$ between these two multinomial distributions, as described below. To account for length differences of $\tilde{W}$ and $\hat{W}$, $\hat{W}$ is first padded by uniform distributions of length $\frac{n}{2}$ on both sides. Then $\tilde{W}$ is slided on positions from leftmost $i=0$ to rightmost $i=m$ on each of the $j$-th filter $\hat{W}^j$, to compute the average distance $d$ over $n$ columns, as measured by Kullback–Leibler divergence. We let the distance between knowledge $\tilde{W}$ and the $j$-th filter $\hat{W^j}$, denoted as $d(\tilde{W}, \hat{W}^j)$, be the minimum distance among all possible positions $i$; and let knowledge dissimilarity function $K$ be the distance of $\tilde{W}$ to the best matched filter $\hat{W}^j$:

\begin{equation}
d(\tilde{W}, \hat{W}^j) = \min( \{D_{KL}(\tilde{W}, \hat{W}^j_{i}), \forall i \} )
\end{equation}
\begin{equation}
K(W,A) = \min( \{d(\tilde{W}, \hat{W}^j), \forall j \} )
\end{equation}

When more than one motif is encoded in knowledge, i.e. $\tilde{W}=\{\tilde{W}^b, b =1,2,.., B\}, B>1$, we treat each $\tilde{W}^b$ independently and compute the arithmetic mean as the final knowledge dissimilarity:
\begin{equation}
K(W,A) = \frac{1}{B} \sum_{b=1}^{B} \min( \{d(\tilde{W}^b, \hat{W}^j), \forall j \} )
\end{equation}

\subsection{Architecture search for joint optimization of loss and knowledge} 

We use reinforcement learning controller network \cite{Zoph2016} as the black-box optimizer in this paper to search neural architectures. The objective function to be optimized is a weighted sum of model loss and model knowledge.

Specifically, we first define a model space of $L$ layers as $\mathbf{\Omega}=\{ \Omega_1, \Omega_2, .., \Omega_L \}$, an ordered collection of different layers. For each layer $\Omega_l, \forall l\in\{1,2,..,L\}$, there are $|\Omega_l|$ number of choices for candidate layer operations and/or hyperparameters, hence the total number of child models in the model space $\mathbf{\Omega}$ is $|\mathbf{\Omega}|=\Pi_{l=1}^{L} |\Omega_l|$. Let $a_t=\{a_{t1}, a_{t2}, .., a_{tL}\}$ be the child model architecture indexed by $t$, where each element $a_{tl} \in \Omega_l$. $a_t$ is a list of one-hot encoded selections for each layer in $\mathbf{\Omega}$, hence fully specifying a child model.

In the given model space $\mathbf{\Omega}$, we seek to find a set of child network architectures $a_t$ that maximizes the reward $R$ on a fixed set of inputs $X$, labels $y$, and knowledge $A$:
\begin{equation}
R(a_t) = - ( L(W; a_t,X,y) + \lambda \cdot K(W,A; a_t) )
\end{equation}
where $L$ is the loss function for the child neural network evaluated on a held-out validation data, $K$ is the knowledge dissimilarity function, $\lambda$ is a non-negative weight for knowledge in optimization, and $W$ is the child model parameters after training.

Since the reward $R$ is non-differentiable, we employ a Recurrent Neural Network, called controller network, parameterized by $\theta$. The controller network generates a child network by sequentially sampling a probability for each layer. For example, in a simple convolutional model space of 4 layers, the controller network predicts convolutional layer, feature pooling, channel pooling and dense layer one by one, where each prediction by a softmax classifier is based on the previous prediction as input. Let $\pi_{\theta}(a_t)$ be the log-likelihood of selecting $a_t$ under the controller with parameters $\theta$; i.e.
\begin{equation}
\pi_{\theta}(a_t) = \sum_{l=1}^{L} \log(P(a_{tl}|a_{t(l-1):1};\theta))
\end{equation}

Intuitively, we would like to update the controller network parameters $\theta$ such that the likelihood $\pi_{\theta}(a_t)$ is increased for generating child networks $a_t$ with high reward, and decreased the likelihood for $a_t$ with low reward. In the original NAS paper \cite{Zoph2016}, the authors followed REINFORCE algorithm to obtain the gradients for updating $\theta$. Let $D$ be a batch of architectures and reward signals, the empirical REINFORCE gradients are computed as
\begin{equation}
\frac{1}{|D|} \sum_{t\in D} \nabla_\theta \pi_{\theta}(a_t) (R_t-b)
\end{equation}
where the baseline function $b$ is an exponential moving average of the previous architecture rewards. In this work, $R_t - b$ is equivalent to the advantage $A_t$, as described below.
A shortcoming of the REINFORCE algorithm is that its gradient updates have high variances. Empirically, performing multiple updates of optimization using REINFORCE gradients often leads to destructively large policy updates \cite{Schulman2017}.

As an alternative, we implemented proximal policy optimization (PPO) \cite{Schulman2017} to train the controller. The PPO policy gradients for controller parameters $\theta$ was empirically obtained through minimizing a surrogate controller loss function $L_{C}$ on the data batch $D$ of architectures and reward signals:
\begin{equation}
L_{C}(\theta) = \frac{1}{|D|} \sum_{t\in D} \min(r_{t}(\theta|\theta_{old})\cdot A_t,  clip(r_t(\theta|\theta_{old}), 1-\epsilon, 1 +\epsilon) \cdot A_t )
\end{equation}
where $\epsilon$ is a hyperparameter for clipping the controller loss function. We followed the previous report \cite{Schulman2017} and set $\epsilon=0.2$. By clipping the surrogate function, PPO prevents destructive policy updates in REINFORCE, and enables multiple epochs of mini-batch updates.

The ratio function $r_{t}(\theta|\theta_{old})$ is the likelihood ratio for selecting $a_t$ under the updated parameters $\theta$ with respect to the old parameters $\theta_{old}$, i.e. 
\begin{equation}
r_{t}(\theta|\theta_{old}) = \exp(\pi_{\theta}(a_t) - \pi_{\theta_{old}}(a_t) )	\end{equation}

The advantage $A_t$ is determined by the current reward $R_t$ subtracting an exponential moving average of previous rewards, $A_t = R_t - EWA(R_{(t-1):1})$. The exponential weight of previous weights is set to 0.8.

\subsection{Simulated data} 

We generated simulated genomic sequence data to benchmark the performance of BioNAS. Using a genomic sequence simulator \cite{simdna-2019}, we first simulated a set of positive DNA sequences ($n=10000$) with a single DNA binding event based on the Transcription Factor (TF) protein motif MYC. Each DNA sequence was set to be 200bp in length. We embedded exactly one motif in each of the positive DNA sequence. Then the negative set ($n=10000$) without any binding event was simulated by random DNA sequences. 

Next we simulated DNA sequences with a mixture of three binding events for transcription factors CTCF, IRF and MYC in $n=100000$ DNA sequences. We set the minimum binding motif occurrence 0, maximum 3, with a mean at 1. Then we also simulated an equal sized set of $n=100000$ negative background sequences without any binding events.

In benchmarking the BioNAS optimization, we defined a bootstrapping procedure to return reward signals from a fixed set of training history of loss and knowledge values. Specifically, for each possible child network architecture $a$ in a model space $\mathbf{\Omega}$, we trained 20 independent models and recorded their loss and knowledge values. The bootstrap function $F_{\Omega}(a_t)$ randomly returns one loss/knowledge pair out of 20 runs for given any $a_t$. In training BioNAS for benchmarking purposes, instead of generating and training the child network independently each time for a new sampled $a_t$, we called the bootstrap function to instantly return the reward signals.

\subsection{ENCODE eCLIP multitask protein binding prediction}

Using the publicly-available eCLIP experiments in ENCODE, we aimed to predict the protein-RNA interaction sites for given RNA sequences. More concretely, for any given RNA sequence, we predict a list of multi-class binary label of whether there are protein binding sites corresponding to a list of proteins of interests. 

All publicly available ENCODE eCLIP peak files were downloaded from the online data portal \cite{Davis2018}. Each gene in the hg19 human genome assembly was divided into 100bp bins, then converted into a vector of binary indicators depending on whether the target bin overlapped with a particular set of peaks. In total, the ENCODE consortium provided $n=333$ datasets, rendering 333 binary labels for each genomic bin. As a result of non-uniform distributed experimental data in the human transcriptome, a negative label in this set might be underpowered for positive signal detection, instead of no true binding sites. To account for this bias, we filtered out bins with less than or equal to 2 peaks across 333 datasets. Then each labeled data was extended symmetrically by 450bp in both directions to consider the genomic context to make each input 1000bp. Finally, all labeled data and inputs from human chromosome 8 were held out for independent testing, while the remaining data were further randomized and split into training and validation with ratio 9:1.

Motif positional weight matrices were compiled from RNAcompete experiments \cite{Ray2013}. All motifs with the corresponding protein data in eCLIP were used as prior knowledge in knowledge dissimilarity function.

\section{Experiments and Results}
\subsection{BioNAS framework for joint optimization of accuracy and knowledge consistency}

We developed BioNAS as an automated machine learning framework to deploy neural network models for bioinformatics applications. In searching the model architectures for a given task, we augmented the loss function with an additional knowledge (dis)similarity function as the optimization objective (Figure \ref{fig1}a). The knowledge function allows flexible incorporation of domain knowledge in searching for neural network architectures. It is defined by transforming and subsequently comparing the existing knowledge and the trained target network parameters in the same space (Methods). In the hypothetical example shown in Figure \ref{fig1}c, two distinct model architectures have comparable predictive powers when optimizing for accuracy (i.e. loss). Using the empirical data alone cannot distinguish the two architectures. Now suppose we have a knowledge function that scores the architectures and trained weights. After considering the knowledge, one architecture compares favorably to the other and hence becomes discriminated by the optimizer. As a proof-of-principle application, we used BioNAS to search convolutional neural network architectures for predicting sequence molecular properties using one-hot encoded genomic sequences as inputs (Figure \ref{fig1}b), augmented with motif-encoding knowledge functions. We applied BioNAS for simulated genomic sequences of single-task and multi-task learning, as well as real functional genomics data from ENCODE eCLIP experiments, as detailed in the following sections.

\begin{figure}[thb]
	\begin{center}
		\includegraphics[scale=0.28]{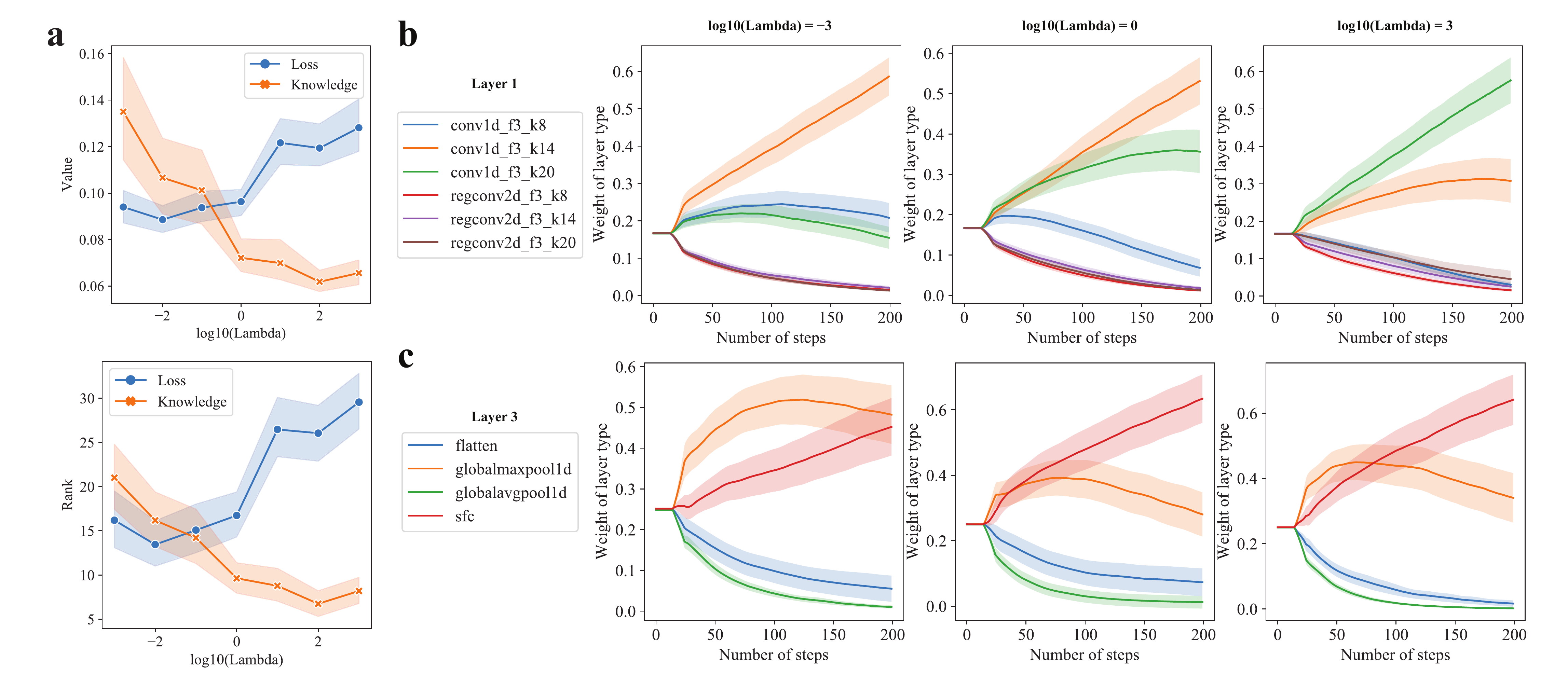}
		\caption{Benchmarking BioNAS with a bootstrapping procedure. {\bf (a)} Gold standard values of loss- and knowledge-dissimilarity functions for converged architecture on varying knowledge weights. {\bf (b)} The rank order of loss- and knowledge-dissimilarity for converged architectures. The total number of architectures is 216; smaller rank is better. {\bf (c)} The average of $n=100$ loss- and knowledge-dissimilarity function values are plotted at each training step of the controller network for knowledge weight $\lambda=0.0001$, and for {\bf (d)} $\lambda=1000$. The values are smoothed by simple moving average; shaded area is the 95\% confidence interval for the plotted average values. }
		\label{fig2}
	\end{center}
\end{figure}

To demonstrate the validity of BioNAS, we first studied a simple 1D-convolutional model for predicting binding sites (bound or not bound) of one transcription factor. Input sequence data was simulated for $n=10000$ positive and negative DNA sequences (Methods). For this simple task, we assumed the underlying true motif was known and subsequently was considered in the knowledge dissimilarity function. We defined a model space of 216 different architectures, each with 4 layers, including convolution, feature-level pooling, channel-level pooling, and dense. Two customized layers that were reported to increase sequence model interpretability were also implemented and/or included \cite{Ploenzke2018}\cite{Alexandari2017}. The detailed model space configuration is shown in Supplementary Table 1.

For a given architecture, the loss and knowledge values are not deterministic, but rather subject to initial weights and stochastic gradient updates. To account for the randomness in parameter initialization and stochastic optimization, we trained 20 models for each architecture with independent initialization and training process, then evaluated its loss and knowledge dissimilarity values. Using these pre-computed loss and knowledge values, BioNAS was trained independently 100 times to find the optimal architecture on each of the varying knowledge weights $\lambda=\{0.001, 0.01, 0.1, 1, 10, 100, 1000\}$ through bootstrapping the pre-computed values (Methods). 

We first show the necessity of incorporating knowledge in neural architecture search. For each architecture, we considered the median value of 20 pre-computed loss and knowledge dissimilarity values as its summary statistic. We compared and ranked the summary statistics of loss and knowledge across all 216 architectures, with smaller ranks representing better architectures. The average of converged architecture summary statistics using different knowledge weights were plotted in Figure \ref{fig2}.
A proper knowledge weight had little or no impact on the predictive power, but dramatically increased the interpretability of converged models ($\lambda \in [0.001, 1]$). However, when the knowledge weight $\lambda$ is too small (i.e. $\lambda \in {0.001, 0.01}$), BioNAS converges to highly predictive models (i.e. low loss), but less interpretable models (i.e. high knowledge dissimilarity), especially when the knowledge weight becomes smaller (e.g. $\lambda=0.001$) (Figure \ref{fig2}a,b). Similarly, when the knowledge weight is too large (i.e. $\lambda \in {10, 100, 1000}$), the converged architectures were more interpretable but less predictive. Overall, models that are both predictive and interpretable can be achieved by BioNAS with a mild knowledge weight (e.g. when $\lambda\in\{0.1,1\}$ in Figure \ref{fig2}a).

Next we examined what types of architectures/layers are contributing to predictive and interpretable models, respectively. We plotted the changes of average weights in the 100 BioNAS runs for each candidate layer type as a function of training steps. As shown in Figure \ref{fig2}b, convolutional layer with kernel size 14 is preferably selected for better predictive power (i.e. $\lambda=0.001$), while convolutional layer with kernel size 20 is selected for better interpretability (i.e. $\lambda=1000$). For $\lambda=1$, a mixture between these two types of layers is observed. The different preferences in convolutional filter size are potentially due to the fact that knowledge function encourages decoding motifs from individual filters, hence a larger filter size is more likely to decode a complete motif. Similarly, separable fully-connected layer works comparably well with global max pooling layer when not considering knowledge dissimilarity, but is more useful for building interpretable models (Figure \ref{fig2}b), a scenario similar to the hypothetical example illustrated in Figure \ref{fig1}b. In the meanwhile, we also observe layers that are consistently essential for both predictive and interpretable models. For example, BioNAS selects dense layer with 10 units over dense 3 units or identity layer (Supplementary Figure 1). Together, we show that a balanced predictive and interpretable model can be achieved by tuning the $\lambda$ parameter and thereby sampling differential target network architectures.

\subsection{Uncovering novel knowledge in multitask learning}

\begin{figure}[htb]
	\begin{center}
		\includegraphics[scale=0.4]{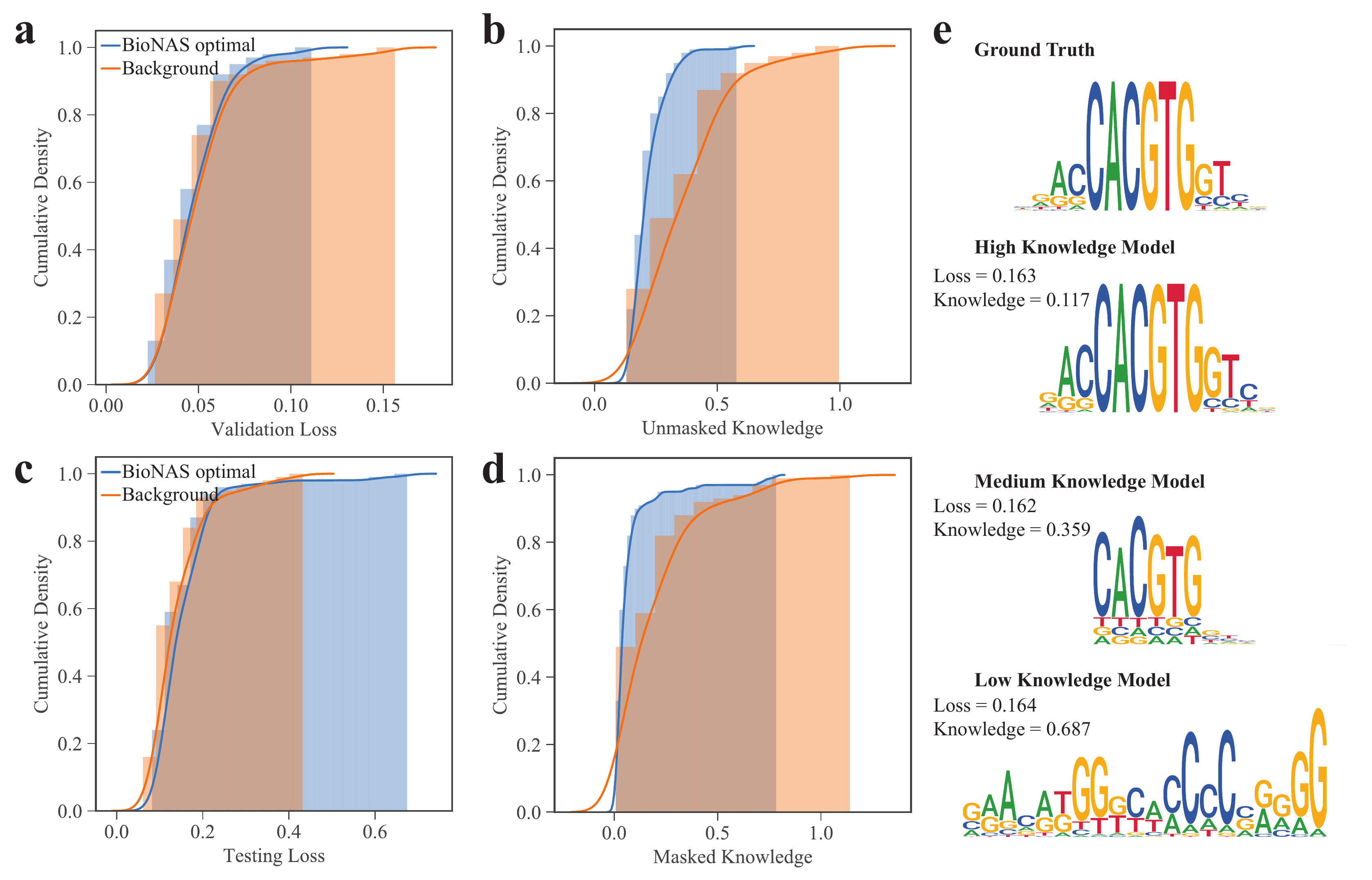}
		\caption{Uncovering novel knowledge by improving model consistency with existing knowledge in multitask learning. A set of background models ($n=100$) were selected to match {\bf (a)} the validation loss distribution of the BioNAS converged optimal models ($n=100$), while {\bf (b)} consistency with the unmasked knowledge (i.e. prior knowledge) was improved in BioNAS optimal models. {\bf (c)} Testing loss distributions and {\bf (d)} Knowledge-dissimilarity distribution for the masked knowledge of BioNAS optimal models and background models. {\bf (e)} Examples of the masked knowledge detection by three models of similar loss but different knowledge-dissimilarity. Top row is the ground truth embedded in the simulation data.}
		\label{fig3}
	\end{center}
\end{figure}

Having demonstrated the optimization behavior on detecting predictive and interpretable architectures, we ask whether we can explicitly learn and extract knowledge from the trained neural network models with BioNAS optimized knowledge consistency. Our hypothesis is that by searching neural network architectures that are in consistency with existing knowledge, we can more reliably decode the trained neural networks and uncover novel knowledge. A distinctive feature in biological and biomedical studies is that the domain researchers have massive prior beliefs and/or knowledge. The knowledge is either derived in the same experimental materials through complementary or orthogonal approaches, or accumulated from external data and information through years of practices and experiences, which provide invaluable guidance in designing the neural network architectures as well as interpreting and understanding the trained model decision logics. BioNAS is designed for flexible incorporation of such domain knowledge (Figure \ref{fig1}).

Multitask learning provides a natural framework for BioNAS to incorporate incomplete existing knowledge and uncover novel knowledge. In multitask learning, multiple relevant sequence properties are jointly learned and predicted with sharing predictive features \cite{Zhou2015}. To emulate a practical multitask learning scenario, we generated a multitask simulation dataset with three TFs (CTCF, IRF, MYC). Suppose two of the three TF binding events (CTCF and IRF) are well-studied, and are therefore considered in the knowledge dissimilarity function as the reward to optimize the neural architecture (i.e. unmasked knowledge). Meanwhile we know nothing about the third TF (MYC), which remains masked. 
A graphical illustration for this multitask learning and the corresponding biological knowledge is in Supplementary Figure 2.

Our goal was to evaluate how well the trained models could recapitulate unmasked knowledge, and more importantly, uncover the masked knowledge. We used $\lambda=1$ to train BioNAS and sample child networks for $n=4$ times independently, in order to account for the randomness in BioNAS training. We defined BioNAS optimal models as a set of models with any one of the four BioNAS-converged target architectures that were sampled and independently trained during training BioNAS. Given the BioNAS-optimal models, we also compiled a set of background models that were not BioNAS-optimal by matching the validation loss distribution ($n=100$; Figure \ref{fig3}a), in order to ensure a fair comparison of knowledge dissimilarities. We further downsampled the BioNAS optimal models to $n=100$ such that both BioNAS-optimal and background groups have the same number of models. For models in both BioNAS-optimal and background sets, they were independently initialized and trained during the training of BioNAS. Indeed, BioNAS-optimal architecture in general has significantly lower knowledge dissimilarity compared to background models (Figure \ref{fig3}b), despite the identical distribution on validation loss.

We next sought to benchmark the model performance on the masked knowledge. As an independent testing set, we simulated $n=1000$ positive sequences by embedding only MYC binding site in the motif, in addition to the training and validation sequences. As expected, the BioNAS-optimal and background models have comparable testing loss distributions on the testing set (Figure \ref{fig3}c), indicating both BioNAS-optimal and background models are comparable in predictive powers. When we evaluated the unmasked knowledge dissimilarity, BioNAS-optimal again outperformed the background models (Figure \ref{fig3}d). This shows that we can more robustly obtain interpretable models to uncover masked/new knowledge, through the optimization of the knowledge dissimilarity on the unmasked/existing knowledge.

As specific examples, we show the sequence logos of masked knowledge decoded from three different trained models (Figure \ref{fig3}e). On the top is the ground-truth motif. 
This motif was not embedded in the knowledge dissimilarity function, and therefore was the knowledge we wished to uncover from the trained models. 
Notably, these three models were very close in testing loss and hence predictive powers. However, the motif decoded from these models were dramatically different. The high knowledge model was one of the BioNAS-converged optimal architectures. It learned a motif highly similar to the ground truth embedded in the training data. By contrast, the medium knowledge model decoded a more divergent motif which misses the flanking patterns; while the low knowledge model could not decode meaningful motifs. The detailed architectures and benchmark metrics are in Supplementary Table 2.

To extend our observations that BioNAS promotes robust discovery of novel knowledge, we repeated the same analysis protocol of MYC, but with masked CTCF and IRF as the knowledge to be uncovered, respectively. Notably, these three motifs were dramatically different (Supplementary Figure 3), hence the uncovered knowledge was not an artifect of overfitting. In both CTCF and IRF analysis, novel knowledge discovery was facilitated if the trained model was more consistent with the existing knowledge. Since the validation loss distributions were matched between BioNAS-optimal and background models, this further demonstrates that knowledge discovery cannot be surrogated by only optimizing for predictive power (left column in Supplementary Figure 3), reinforcing the importance of considering knowledge in searching for neural architectures.

\subsection{Multitask architecture search for protein-RNA binding site prediction using ENCODE eCLIP }

\begin{figure}[htb]
	\begin{center}
		\includegraphics[scale=0.4]{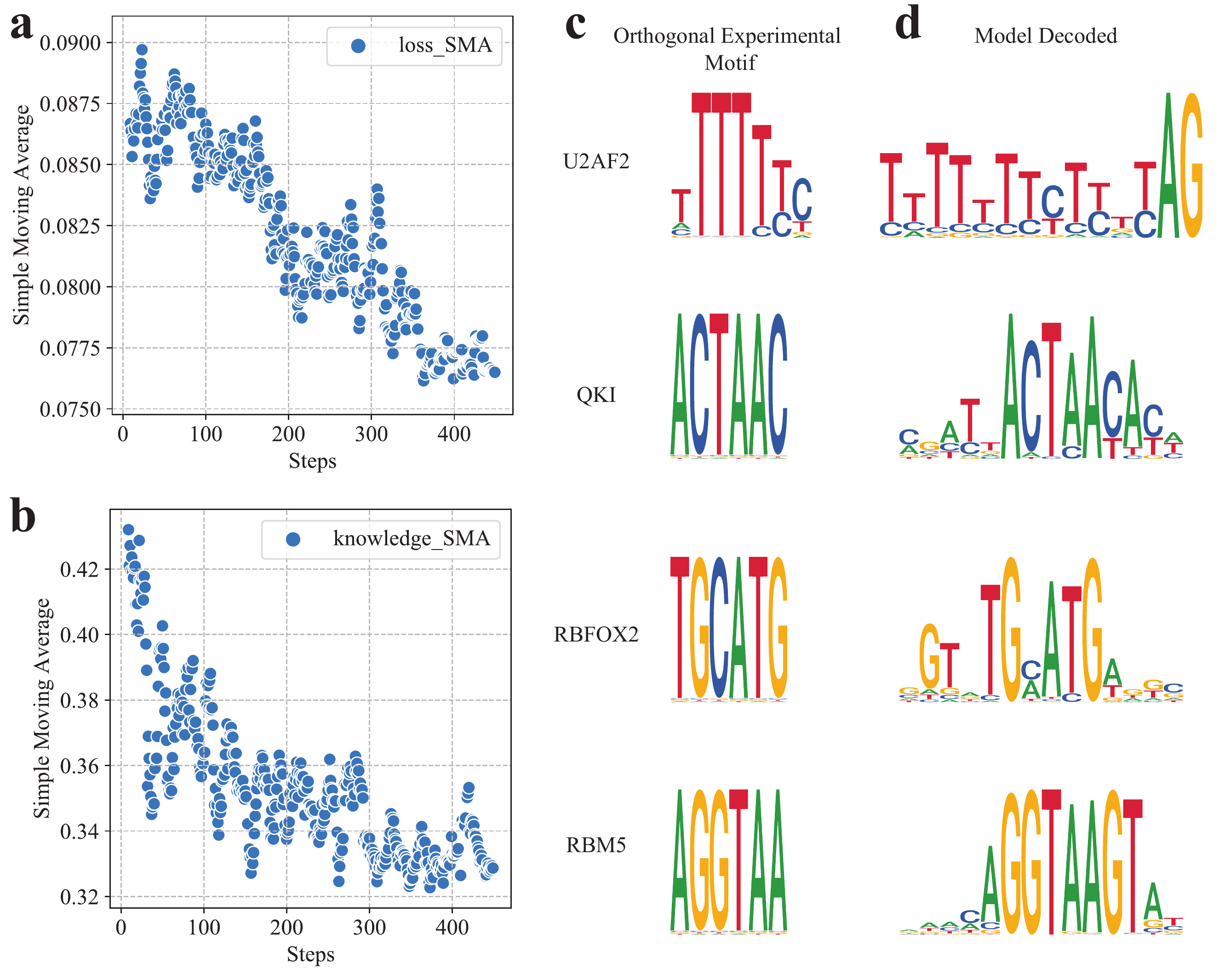}
		\caption{BioNAS application in ENCODE eCLIP multitask learning. The {\bf (a)} loss and {\bf (b)} knowledge dissimilarity function values were optimized as the controller network generated model architectures for eCLIP binding prediction.  {\bf (c)} Existing prior knowledge from orthogonal experiments and {\bf (d)} the model decoded motifs were visualized.}
		\label{fig4}
	\end{center}
\end{figure}

Finally, we applied BioNAS to study the protein-RNA binding sites using the ENCODE eCLIP data \cite{VanNostrand2016}. Briefly, for any given RNA sequence, the task is to predict a multi-class binary outcome (bound or not bound) corresponding to a set of proteins binding to the RNA molecule with that sequence.
eCLIP and its variant CLIP experiments are immunoprecipitation-based experimental approaches to identify the binding events of a target RNA-binding protein (RBP) across the transcriptome \cite{VanNostrand2016}. Existing computational methods find putative protein binding sites across the transcriptome by a statistical procedure called peak calling using CLIP data \cite{Zhang2017}. The CLIP technology provides a global landscape of the protein-RNA binding events in living tissue or cells. Therefore, the readout of CLIP experiments reflect not only the comprehensive binding patterns for the protein of interest, but also the potential combinatorial interactions and tissue-specific regulation with other genes.

Meanwhile, a complementary experimental approach to study the RBP binding patterns has been RNAcompete \cite{Ray2013}, which is directly enriching short synthesized RNA oligo-nucleotides bound by the RBP of interest in order to determine its sequence preferences, and is more easily interpretable compared to eCLIP. The shortcoming of RNAcompete experiment is that it loses any potential context information (e.g. flanking motifs of interacting protein partners) because of being conducted outside living cells. The knowledge provided from RNAcompete experiments is explicit but potentially incomplete.

Using BioNAS, our aim is to generate more interpretable models for predicting eCLIP binding sites by leveraging the explicit prior knowledge in the orthogonal RNAcompete experiments. More concretely, the RNAcompete information provides guidance to decode and annotate the convolutional feature maps in the multitask learning eCLIP model. In return, the interpretable eCLIP model can uncover novel knowledge in living cells to supplement the incomplete RNAcompete knowledge outside living cells.

We first compiled the training labels and genomic sequences from ENCODE eCLIP data (Methods). In total we had $n=538955$ sequences for training and $n=59883$ for validation (Methods). Then a large model space with 9 layers and in total 1,166,400 possible architectures was compiled for BioNAS to search for target model architectures (Supplementary Table 3). As the training of controller progressed and the number of generated models increased, the loss and knowledge dissimilarity values decreased  (Figure \ref{fig4}a,b), indicating the controller was generating better child models for both predictive power and interpretability over time. Using the orthogonal RNAcompete experiment knowledge, we decoded the BioNAS-optimal model trained using eCLIP into corresponding motifs that had small knowledge dissimilarities to existing RNAcompete motifs. Four of the top model decoded motifs and their matching RNAcompete motifs were visualized in Figure \ref{fig4}c,d. The feature maps of these convolutional filters trained from eCLIP data were highly similar to RNAcompete motifs of known RBP genes, therefore could be interpreted and annotated to specific genes, facilitating downstream analyses and interpretation of the decision logics in the trained eCLIP model \cite{ZhangQ2017}.

Of particular interest, the decoded model weights of U2AF2 not only matched the known motif, but also captured the 3'-splice site pattern AG (top row; Figure \ref{fig4}c,d). The AG splice site is a well-established pattern for RNA splicing, and the binding of U2AF2 upstream to 3'-splice sites is essential for splicing to take place \cite{Fu2014}. Biologically, U2AF2 is part of the core spliceosome constituted of several subunit genes in living cells; in particular, its interacting partner U2AF1 recognizes and binds to the 3'-splice sites. Together with other RBPs, U2AF1/2 maintains the fidelity of RNA splicing. This splice site pattern was not present in the prior RNAcompete knowledge due to the \textit{in vitro} nature of the RNAcompete experiment - the lack of U2AF2's interacting partners outside living cells. Within living cells, despite the fact that there are many sequence instances across the genome that satisfy this \textit{in vitro} motif, only a small fraction of them near the 3'-splice site are \textit{bona fide} binding sites of U2AF2. 
Utilizing the existing partial knowledge, the BioNAS-optimal model uncovered this full motif plus splice-site pattern from the eCLIP data, which otherwise will not be annotated without RNAcompete data nor discovered without eCLIP data. Similarly, for proteins QKI, RBFOX2 and RBM5, the model decoded motif matched the partial knowledge, while also uncovered additional flanking patterns from eCLIP (Figure \ref{fig4}c,d). These patterns potentially represented additional sequence preferences of the specific RBP genes and/or their interacting partner genes; however, the detailed experimental and functional validations for the flanking motif patterns were beyond the scope of this work.

Collectively, we demonstrated how BioNAS optimized the knowledge consistency to existing RNAcompete knowledge and led to the discovery of novel knowledge in the eCLIP models. By augmenting eCLIP CNN model training with information from RNAcompete, we annotated the convolutional feature maps with biological semantics. These interpretable feature maps potentially shed new lights on the regulatory patterns of RBP genes in living cells.

\section{Conclusion}
Opening the black box of deep learning applications is a crucial yet non-trivial challenge in biomedicine and genomics. We propose to answer this challenge by searching the model architecture space for joint optimization of predictive power and biological knowledge. The introduction of knowledge functions that measure the dissimilarity between existing knowledge with the decoded trained model knowledge is powered by ongoing efforts in two fields: in life sciences, knowledge encoders are developed for encoding experimental results; in machine learning, model decoders are developed for understanding the trained model decision rules. 
Using our proposed framework BioNAS, domain experts can substantially reduce the computational barrier to access deep learning, and easily find deep learning models that are both high predictive and interpretable for their specific data and applications. Furthermore, BioNAS uncovers novel knowledge by optimizing the model consistency to the existing knowledge, as demonstrated in both simulated and real data of functional genomics. 

With the continuing accumulation of genomics and biomedical data, deep learning-based methods provide an essential workhorse for various tasks. We anticipate BioNAS will be a valuable contribution in adopting deep learning to the life science research communities.

\bibliography{main}

\begin{thebibliography}{10}

\bibitem{Alexandari2017}
Amr~Mohamed Alexandari, Avanti Shrikumar, and Anshul Kundaje.
\newblock {Separable Fully Connected Layers Improve Deep Learning Models For
  Genomics}.
\newblock {\em bioRxiv}, page 146431, jul 2017.

\bibitem{Alipanahi2015}
Babak Alipanahi, Andrew Delong, Matthew~T Weirauch, and Brendan~J Frey.
\newblock {Predicting the sequence specificities of DNA- and RNA-binding
  proteins by deep learning}.
\newblock {\em Nature Biotechnology}, 33(8):831--838, jul 2015.

\bibitem{Bailey1994}
T~L Bailey and C~Elkan.
\newblock {Fitting a mixture model by expectation maximization to discover
  motifs in biopolymers.}
\newblock {\em Proceedings. International Conference on Intelligent Systems for
  Molecular Biology}, 2:28--36, 1994.

\bibitem{Ching2018}
Travers Ching, Daniel~S Himmelstein, Brett~K Beaulieu-Jones, Alexandr~A
  Kalinin, Brian~T Do, Gregory~P Way, Enrico Ferrero, Paul-Michael Agapow,
  Michael Zietz, Michael~M Hoffman, Wei Xie, Gail~L Rosen, Benjamin~J
  Lengerich, Johnny Israeli, Jack Lanchantin, Stephen Woloszynek, Anne~E
  Carpenter, Avanti Shrikumar, Jinbo Xu, Evan~M Cofer, Christopher~A Lavender,
  Srinivas~C Turaga, Amr~M Alexandari, Zhiyong Lu, David~J Harris, Dave
  DeCaprio, Yanjun Qi, Anshul Kundaje, Yifan Peng, Laura~K Wiley, Marwin H~S
  Segler, Simina~M Boca, S~Joshua Swamidass, Austin Huang, Anthony Gitter, and
  Casey~S Greene.
\newblock {Opportunities and obstacles for deep learning in biology and
  medicine.}
\newblock {\em Journal of the Royal Society, Interface}, 15(141):20170387, apr
  2018.

\bibitem{Davis2018}
Carrie~A Davis, Benjamin~C Hitz, Cricket~A Sloan, Esther~T Chan, Jean~M
  Davidson, Idan Gabdank, Jason~A Hilton, Kriti Jain, Ulugbek~K Baymuradov,
  Aditi~K Narayanan, Kathrina~C Onate, Keenan Graham, Stuart~R Miyasato,
  Timothy~R Dreszer, J~Seth Strattan, Otto Jolanki, Forrest~Y Tanaka, and
  J~Michael Cherry.
\newblock {The Encyclopedia of DNA elements (ENCODE): data portal update.}
\newblock {\em Nucleic acids research}, 46(D1):D794--D801, jan 2018.

\bibitem{Dhaeseleer2006}
Patrik D'haeseleer.
\newblock {What are DNA sequence motifs?}
\newblock {\em Nature Biotechnology}, 24(4):423--425, apr 2006.

\bibitem{Eraslan2019}
G{\"{o}}kcen Eraslan, {\v{Z}}iga Avsec, Julien Gagneur, and Fabian~J. Theis.
\newblock {Deep learning: new computational modelling techniques for genomics}.
\newblock {\em Nature Reviews Genetics}, page~1, apr 2019.

\bibitem{Esteva2017}
Andre Esteva, Brett Kuprel, Roberto~A. Novoa, Justin Ko, Susan~M. Swetter,
  Helen~M. Blau, and Sebastian Thrun.
\newblock {Dermatologist-level classification of skin cancer with deep neural
  networks}.
\newblock {\em Nature}, 542(7639):115--118, feb 2017.

\bibitem{Fu2014}
Xiang-Dong Fu and Manuel Ares.
\newblock {Context-dependent control of alternative splicing by RNA-binding
  proteins}.
\newblock {\em Nature Reviews Genetics}, 15(10):689--701, aug 2014.

\bibitem{Kandasamy2019}
Kirthevasan Kandasamy, Karun~Raju Vysyaraju, Willie Neiswanger, Biswajit Paria,
  Christopher~R. Collins, Jeff Schneider, Barnabas Poczos, and Eric~P. Xing.
\newblock {Tuning Hyperparameters without Grad Students: Scalable and Robust
  Bayesian Optimisation with Dragonfly}.
\newblock mar 2019.

\bibitem{Kelley2016}
David~R Kelley, Jasper Snoek, and John~L Rinn.
\newblock {Basset: learning the regulatory code of the accessible genome with
  deep convolutional neural networks.}
\newblock {\em Genome research}, 26(7):990--9, may 2016.

\bibitem{simdna-2019}
kundaje lab.
\newblock simdna: simulated datasets of dna,.
\newblock \url{https://github.com/kundajelab/simdna}, 2019.

\bibitem{smac-2017}
Marius Lindauer, Katharina Eggensperger, Matthias Feurer, Stefan Falkner,
  André Biedenkapp, and Frank Hutter.
\newblock Smac v3: Algorithm configuration in python.
\newblock \url{https://github.com/automl/SMAC3}, 2017.

\bibitem{Liu2018}
Hanxiao Liu, Karen Simonyan, and Yiming Yang.
\newblock {DARTS: Differentiable Architecture Search}.
\newblock jun 2018.

\bibitem{Ploenzke2018}
MS~Ploenzke and RA~Irizarry.
\newblock {Interpretable Convolution Methods for Learning Genomic Sequence
  Motifs}.
\newblock {\em bioRxiv}, page 411934, sep 2018.

\bibitem{Ray2013}
Debashish Ray, Hilal Kazan, Kate~B. Cook, Matthew~T. Weirauch, Hamed~S.
  Najafabadi, Xiao Li, Serge Gueroussov, Mihai Albu, Hong Zheng, Ally Yang,
  Hong Na, Manuel Irimia, Leah~H. Matzat, Ryan~K. Dale, Sarah~A. Smith,
  Christopher~A. Yarosh, Seth~M. Kelly, Behnam Nabet, Desirea Mecenas, Weimin
  Li, Rakesh~S. Laishram, Mei Qiao, Howard~D. Lipshitz, Fabio Piano, Anita~H.
  Corbett, Russ~P. Carstens, Brendan~J. Frey, Richard~A. Anderson, Kristen~W.
  Lynch, Luiz O.~F. Penalva, Elissa~P. Lei, Andrew~G. Fraser, Benjamin~J.
  Blencowe, Quaid~D. Morris, and Timothy~R. Hughes.
\newblock {A compendium of RNA-binding motifs for decoding gene regulation}.
\newblock {\em Nature}, 499(7457):172, 2013.

\bibitem{Ribeiro2016}
Marco~Tulio Ribeiro, Sameer Singh, and Carlos Guestrin.
\newblock {"Why Should I Trust You?": Explaining the Predictions of Any
  Classifier}.
\newblock feb 2016.

\bibitem{Schulman2017}
John Schulman, Filip Wolski, Prafulla Dhariwal, Alec Radford, and Oleg Klimov.
\newblock {Proximal Policy Optimization Algorithms}.
\newblock jul 2017.

\bibitem{Shrikumar2017}
Avanti Shrikumar, Peyton Greenside, and Anshul Kundaje.
\newblock {Learning Important Features Through Propagating Activation
  Differences}.
\newblock apr 2017.

\bibitem{VanNostrand2016}
Eric~L {Van Nostrand}, Gabriel~A Pratt, Alexander~A Shishkin, Chelsea
  Gelboin-Burkhart, Mark~Y Fang, Balaji Sundararaman, Steven~M Blue, Thai~B
  Nguyen, Christine Surka, Keri Elkins, Rebecca Stanton, Frank Rigo, Mitchell
  Guttman, and Gene~W Yeo.
\newblock {Robust transcriptome-wide discovery of RNA-binding protein binding
  sites with enhanced CLIP (eCLIP)}.
\newblock {\em Nature Methods}, 13(6):508--514, mar 2016.

\bibitem{Xiong2015}
Hui~Y Xiong, Babak Alipanahi, Leo~J Lee, Hannes Bretschneider, Daniele Merico,
  Ryan K~C Yuen, Yimin Hua, Serge Gueroussov, Hamed~S Najafabadi, Timothy~R
  Hughes, Quaid Morris, Yoseph Barash, Adrian~R Krainer, Nebojsa Jojic,
  Stephen~W Scherer, Benjamin~J Blencowe, and Brendan~J Frey.
\newblock {RNA splicing. The human splicing code reveals new insights into the
  genetic determinants of disease.}
\newblock {\em Science (New York, N.Y.)}, 347(6218):1254806, jan 2015.

\bibitem{Yu2018}
Michael~K. Yu, Jianzhu Ma, Jasmin Fisher, Jason~F. Kreisberg, Benjamin~J.
  Raphael, and Trey Ideker.
\newblock {Visible Machine Learning for Biomedicine}.
\newblock {\em Cell}, 173(7):1562--1565, jun 2018.

\bibitem{ZhangQ2017}
Quanshi Zhang, Ying~Nian Wu, and Song-Chun Zhu.
\newblock {Interpretable Convolutional Neural Networks}.
\newblock oct 2017.

\bibitem{Zhang2019}
Zijun Zhang, Zhicheng Pan, Yi~Ying, Zhijie Xie, Samir Adhikari, John Phillips,
  Russ~P. Carstens, Douglas~L. Black, Yingnian Wu, and Yi~Xing.
\newblock {Deep-learning augmented RNA-seq analysis of transcript splicing}.
\newblock {\em Nature Methods 2019 16:4}, 16(4):307, mar 2019.

\bibitem{Zhang2017}
Zijun Zhang and Yi~Xing.
\newblock {CLIP-seq analysis of multi-mapped reads discovers novel functional
  RNA regulatory sites in the human transcriptome}.
\newblock {\em Nucleic Acids Research}, 45(16):9260--9271, sep 2017.

\bibitem{Zhou2019}
Jian Zhou, Christopher~Y. Park, Chandra~L. Theesfeld, Aaron~K. Wong, Yuan Yuan,
  Claudia Scheckel, John~J. Fak, Julien Funk, Kevin Yao, Yoko Tajima, Alan
  Packer, Robert~B. Darnell, and Olga~G. Troyanskaya.
\newblock {Whole-genome deep-learning analysis identifies contribution of
  noncoding mutations to autism risk}.
\newblock {\em Nature Genetics}, 51(6):973--980, jun 2019.

\bibitem{Zhou2015}
Jian Zhou and Olga~G Troyanskaya.
\newblock {Predicting effects of noncoding variants with deep learning–based
  sequence model}.
\newblock {\em Nature Methods}, 12(10):931--934, oct 2015.

\bibitem{Zhou2004}
Q.~Zhou and W.~H. Wong.
\newblock {CisModule: De novo discovery of cis-regulatory modules by
  hierarchical mixture modeling}.
\newblock {\em Proceedings of the National Academy of Sciences},
  101(33):12114--12119, aug 2004.

\bibitem{Zoph2016}
Barret Zoph and Quoc~V. Le.
\newblock {Neural Architecture Search with Reinforcement Learning}.
\newblock nov 2016.

\end{thebibliography}
\bibliographystyle{plain}

\end{document}


{\bf Supplementary Information}

\begin{figure}[htb]
	\begin{center}
		\includegraphics[scale=0.4]{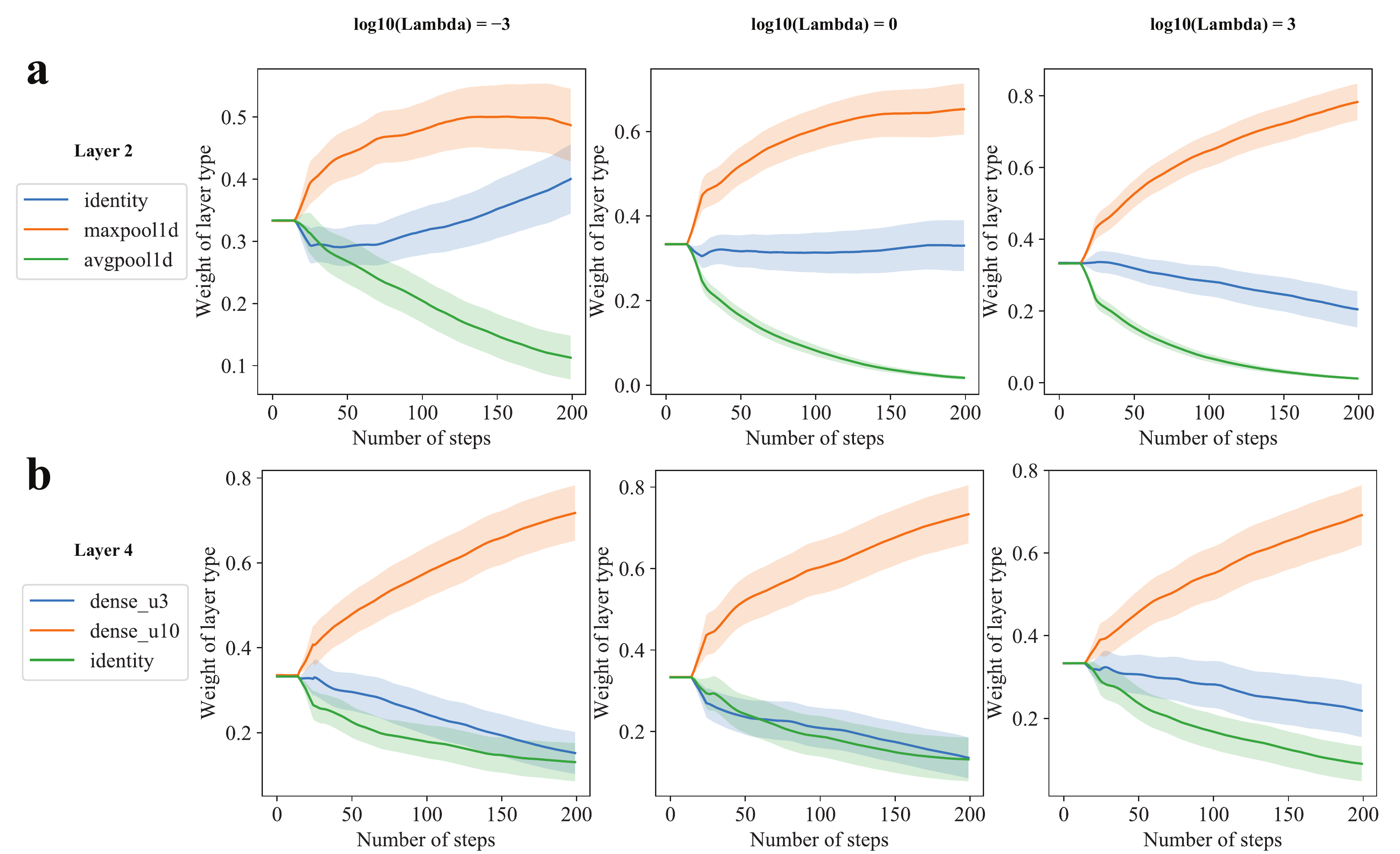}
		\caption{Layers consistently selected by BioNAS on different knowledge weight values. Max pooling is consistently selected on Layer 2 and Dense unit 10 layer is selected on Layer 4.}
		\label{supp_fig1}
	\end{center}
\end{figure}

\begin{figure}[htb]
	\begin{center}
		\includegraphics[scale=0.7]{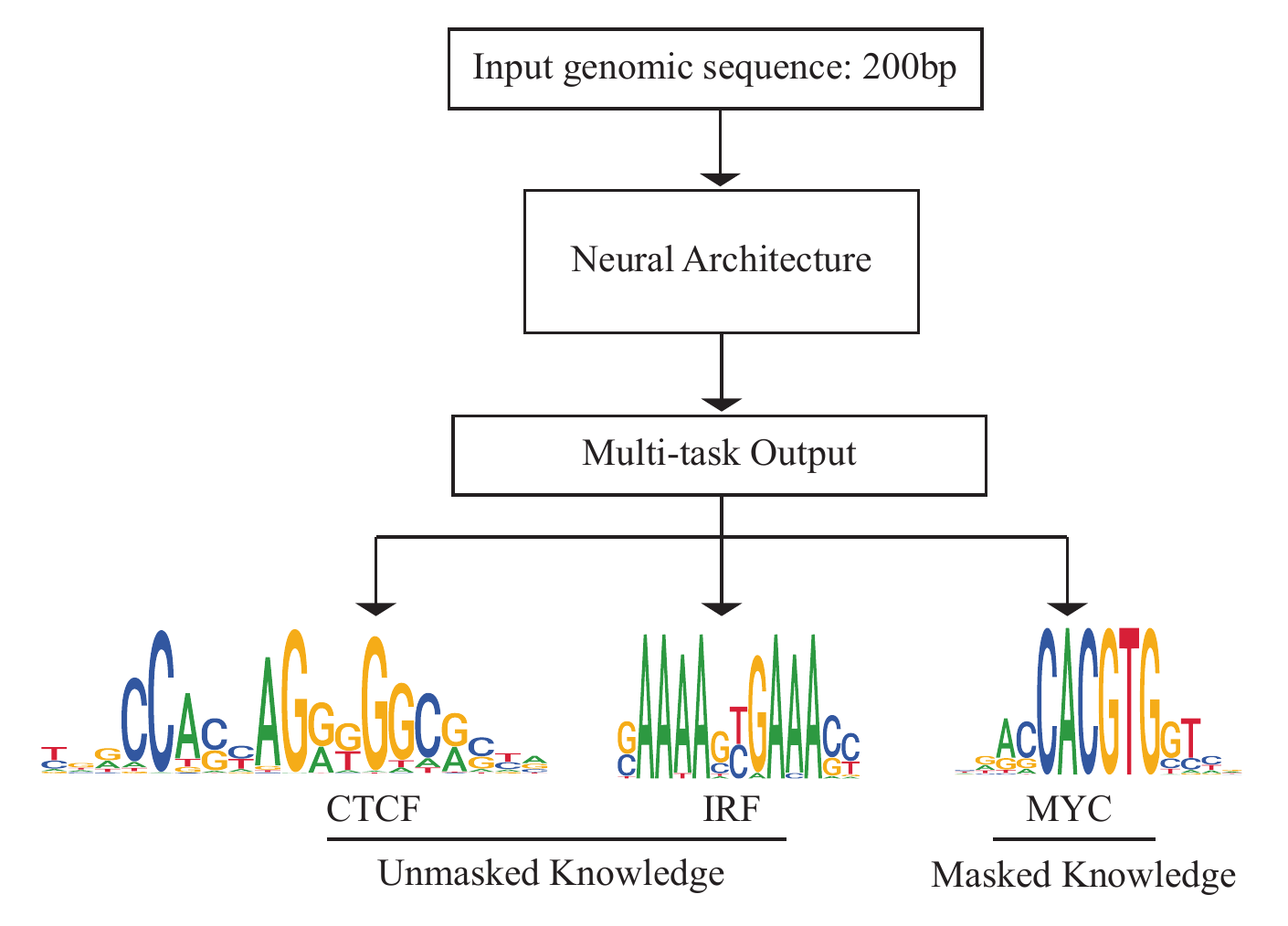}
		\caption{An overview of the multi-tasking TF prediction task. Two TFs, CTCF and IRF, are unmasked knowledge to optimize the target network architecture. The third TF MYC is masked for evaluation purpose.}
		\label{supp_fig2}
	\end{center}
\end{figure}

\begin{figure}[htb]
	\begin{center}
		\includegraphics[scale=0.4]{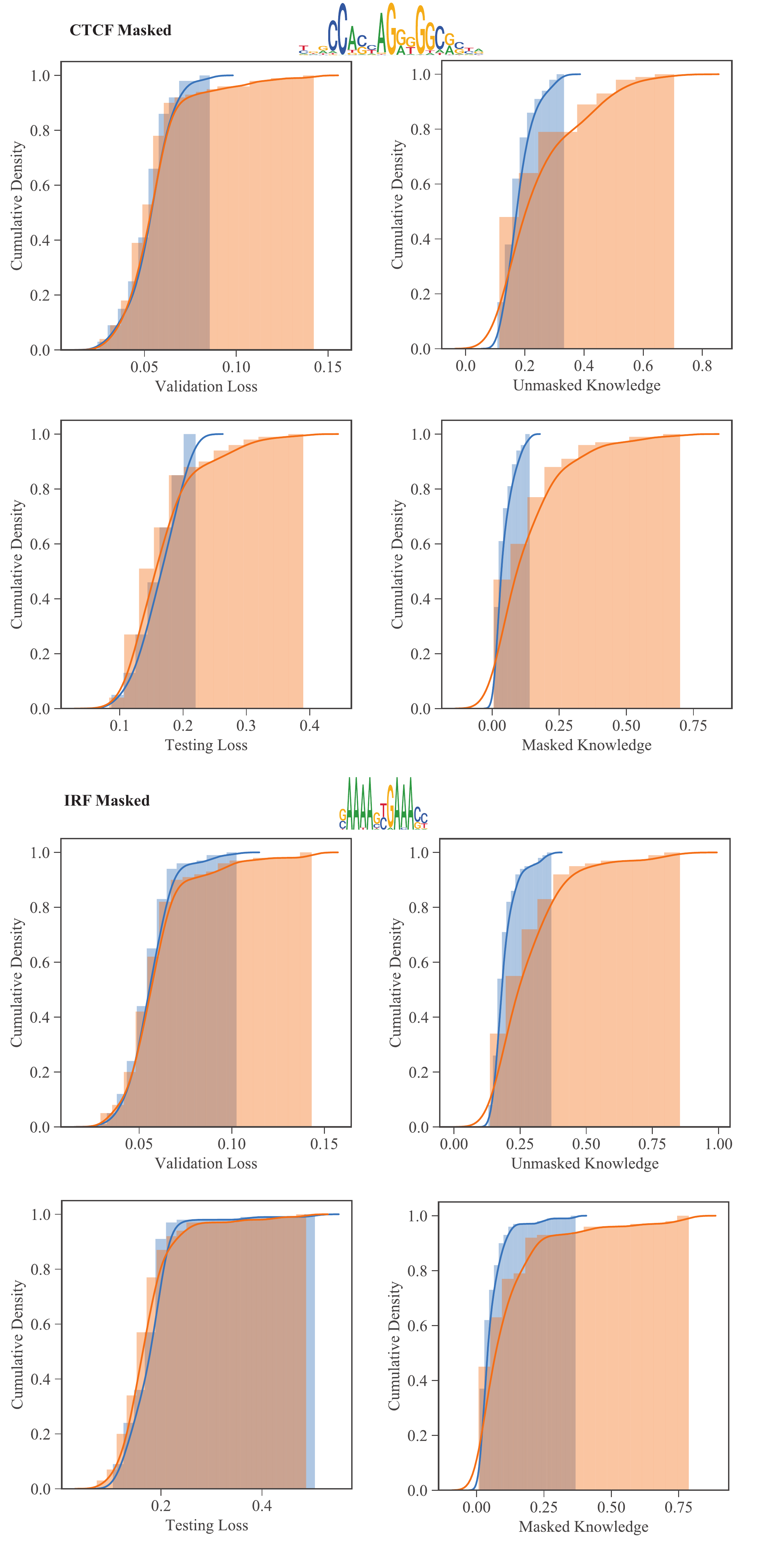}
		\caption{BioNAS performance when masking CTCF and IRF in the knowledge function. The loss and knowledge distributions were analyzed and plotted following the identical steps as in Figure 4.}
		\label{supp_fig3}
	\end{center}
\end{figure}

\begin{table}[htb]
	\caption{Model space for a simple 1D-convolutional task.}
	\label{supp_table1}
	\centering
	\begin{tabular}{llc}
		\hline
		& \multicolumn{1}{c}{\textbf{Candidate Layers}}                                                                                                          & \textbf{No. of Choices} \\ \hline
		Layer 1 & \begin{tabular}[c]{@{}l@{}}conv1d\_f3\_k8, conv1d\_f3\_k14, conv1d\_f3\_k20, \\ regconv2d\_f3\_k8, regconv2d\_f3\_k14, regconv2d\_f3\_k20\end{tabular} & 6                       \\ 
		Layer 2 & identity, maxpool1d, avgpool1d                                                                                                                         & 3                       \\ 
		Layer 3 & flatten, globalmaxpool1d, globalavgpool1d, sfc                                                                                                         & 4                       \\ 
		Layer 4 & dense\_u3, dense\_u10, identity                                                                                                                        & 3                       \\ \hline
		\textbf{Total} & & 216 \\ \hline
	\end{tabular}
\end{table}

\begin{sidewaystable}
	\centering
	\caption{Detailed information for three representative models on multitasking TF prediction}
	\label{supp_table2}
	\begin{tabular}{@{}lllllllllll@{}}
		\toprule
		\textbf{Name} & \textbf{ID} & \textbf{Layer1}         & \textbf{Layer2} & \textbf{Layer3}     & \textbf{Layer4} & \textbf{acc} & \textbf{Unmasked Knowledge} & \textbf{Val Loss} & \textbf{Masked Knowledge} & \textbf{Test Loss} \\ \midrule
		High          & 15          & conv1d\_f10\_k20    & maxpool1d   & globalmaxpool1d & dense\_u10  & 0.983  & 0.173                  & 0.061        & 0.117                & 0.163          \\
		Medium        & 97          & regconv2d\_f10\_k14 & maxpool1d   & globalmaxpool1d & dense\_u30  & 0.987   & 0.426                  & 0.050        & 0.359                & 0.162          \\
		Low           & 90          & conv1d\_f10\_k20    & avgpool1d   & globalavgpool1d & dense\_u30  & 0.984  & 0.731                  & 0.0569        & 0.687                & 0.164          \\ \bottomrule
	\end{tabular}
\end{sidewaystable}

\begin{table}[htb]
	\caption{Model space for searching ENCODE eCLIP multi-tasking model architectures.}
	\label{supp_table3}
	\begin{tabular}{@{}llll@{}}
		\toprule
		& \textbf{Layer Type} & \textbf{Candidate Layers}                                 & \textbf{No. of choices} \\ \midrule
		Layer 1        & conv1d              & filters=\{100,300,500\};kernel\_size=\{20,14,8\}          & 9                       \\
		Layer 2        & feature-pooling     & identity; maxpool1d; avgpool1d                            & 3                       \\
		Layer 3        & conv1d/identity     & identity|filters=\{100,300,500\};kernel\_size=\{20,14,8\} & 10                      \\
		Layer 4        & feature-pooling     & identity; maxpool1d; avgpool1d                            & 3                       \\
		Layer 5        & conv1d/identity     & identity|filters=\{100,300,500\};kernel\_size=\{20,14,8\} & 10                      \\
		Layer 6        & feature-pooling     & identity; maxpool1d; avgpool1d                            & 3                       \\
		Layer 7        & channel-pooling     & flatten; globalmaxpool; globalavgpool; sfc                & 4                       \\
		Layer 8        & regularization      & identity; sparsek; dropout                                & 3                       \\
		Layer 9        & dense/identity      & identity|units=\{100,300,500\}                            & 4                       \\ \hline
		\textbf{Total} &                     &                                                           & \textbf{1,166,400}        \\ \bottomrule
	\end{tabular}
\end{table}